\documentclass[lettersize,journal]{IEEEtran}
\usepackage{amsmath,amsfonts}
\usepackage{algorithmic}
\usepackage{algorithm}
\usepackage{array}
\usepackage{textcomp}
\usepackage{stfloats}
\usepackage{url}
\usepackage{verbatim}
\usepackage{graphicx}
\usepackage{cite}
\usepackage{booktabs}
\usepackage{caption} 
\usepackage{subfigure} 

\usepackage{bm}
\usepackage{amssymb}
\usepackage{xcolor}
\usepackage{balance}
\usepackage{multirow} 
\usepackage[numbers,sort&compress]{natbib}

\begin{document}

\title{A Simple yet Effective Subway Self-positioning Method based on Aerial-view Sleeper Detection}
\author{Jiajie Song, Ningfang Song, Xiong Pan, Xiaoxin Liu, Can Chen, and Jingchun Cheng*
\thanks{*Corresponding Author: Jingchun Cheng, chengjingchun14@163.com}
\thanks{The authors are affiliated with the School of  Instrumentation and Optoelectronic Engineering, Beihang University, Beijing 100080, China.}
}

\maketitle
\begin{abstract}
With the rapid development of urban underground rail vehicles,
subway positioning,
which plays a fundamental role in the traffic navigation and collision avoidance systems,
has become a research hot-spot these years.
Most current subway positioning methods rely on localization beacons densely pre-installed alongside the railway tracks, requiring massive costs for infrastructure and maintenance, while commonly lacking flexibility and anti-interference ability.
In this paper, we propose a low-cost and real-time visual-assisted self-localization framework to address the robust and convenient positioning problem for subways. 
Firstly, we perform aerial view rail sleeper detection based on the fast and efficient YOLOv8n network. 
The detection results are then used to achieve real-time correction of mileage values combined with geometric positioning information, obtaining precise subway locations. 
Front camera Videos for subway driving scenes along a 6.9 km
route are collected and annotated from the simulator for validation of the proposed method.
Experimental results show that our aerial view sleeper detection algorithm can efficiently detect sleeper positions with F1-score of 0.929 at 1111 fps,
and that the proposed positioning framework achieves a mean percentage error of 0.1\%,
demonstrating its continuous and high-precision self-localization capability.
\end{abstract}

\begin{IEEEkeywords}
Subway positioning, vision-based self-positioning, visual odometer, deep Learning.
\end{IEEEkeywords}

\section{Introduction}
\IEEEPARstart{S}{ubways} are gradually becoming a main component of the urban transportation system, as they offer prominent advantages such as economizing floor space, saving time, ensuring punctuality, accommodating large passenger capacity, and more.
An indispensable part of keeping the smooth running of subway trains is the subway positioning system, 
which can provide positioning information for subway trains in the complex underground tunnel and station environments. 
The accuracy of the subway positioning system directly affects the tracking, navigation, and intelligent scheduling of subway trains \cite{R1}; it further has an influence on the safety, efficiency, and maintenance of the whole underground transportation system.
As a result, the requirement for advanced, reliable, and stable underground positioning techniques has become a significant and long-lasting practical demand.

In this paper, we tackle the problem of subway positioning, which aims to provide the real-time, accurate positioning information continuously and stably along the entire driving route.
However, in practice, the subway positioning task is far more challenging than the positioning of cars or overground trains.
This is because ground vehicles have the superiority of accessing the satellite positioning systems (GPS, BDS, etc.),
allowing them to obtain the accurate longitude and latitude information to eliminate accumulated errors caused by inertial sensors. \cite{R2, R3, R4, R5}.
For subways whose routes are mostly underground, the mature vehicle positioning systems relying on satellite signals lose efficacy due to signal masking.
As a result, subway positioning systems are forced to rely on high-cost hardware equipment to bring satellite signals underground,
or have to introduce new reference locating signals \cite{R6, R7, R8, R9, R10, R11}.
Most currently-in-use subway positioning methods are based on track-side localization beacons \cite{R6, R8, R51, R52} like the balise devices
that send telegrams containing accurate position information to passing-by subways.
With the help of on-board sensors like Inertial Measurement Units (IMU) and wheel speed odometers, subways can expand the discrete position ground-truths (provided by beacons) to continuous positions by interpolation or other techniques.
However, such positioning systems require a large number of pre-installed localization beacons along the whole railroad, resulting in vast amounts of construction and maintenance costs \cite{R11}.
Besides, under emergency conditions, track-side beacons may fail to send telegrams with their working conditions affected by the unstable railroad voltages \cite{R10}.
Therefore, from the point of cost and safety, it is instructive and meaningful to study the subway self-positioning task, where subways can obtain self-positions directly from on-board sensors with no outside intervention.  

As introduced above, subway self-positioning requires methods to achieve active positioning of subways with only the on-board sensors like cameras, LIDARs, radars, IMUs etc.
As radars are of limited detection zone and IMUs have over-time accumulation errors, 
LIDARs and cameras often play the role of main information source for self-positioning methods, where LIDARs can provide fine-grained 3D point cloud information via active scanning, and cameras can provide high-resolution visual information at low cost.
For example, with the high-precision LIDAR sensors, \cite{R12} combines the Renyi’s Quadratic Entropy (RQE) based point cloud alignment algorithm with sliding-window-filtered odometer to form a railroad self-positioning framework; it achieves satisfactory positioning accuracy, but is difficult to popularize due to the high-cost and vulnerability of LIDAR sensors.
In contrast,
cameras can steadily provide high-resolution, large-coverage, rich, and detailed visual information at low cost.
Equipping with accurate intelligent positioning algorithms, visual-based subway self-positioning systems have more extensive spreading values and prospects for application.
For example, \cite{R37} develops a low-cost camera-based subway self-positioning method with track and switch recognition.
The system runs with large-coverage, rich, and detailed visual information provided by a wide-angle lens installed on the train coupler at a rather low cost, 
but has lower accuracy than \cite{R12} 
due to the sensitivity and information dimension variance between the two types of sensing data. 
Recently, many high-precision visual positioning methods are developed based on emerging deep learning techniques, like visual odometry (VO) \cite{R13}, visual-initial odometry (VIO) \cite{R14}, simultaneous localization and mapping (SLAM) \cite{R15} methods, and etc.
These methods have shown amazing performance in various localization tasks \cite{R16, R17, R18}, and can help researchers narrow the performance gap between camera and LIDAR in the subway elf-positioning task.
For instance, \cite{R19} investigates various types of vision-assisted odometry information for train localization and proposes an optimal combination for simultaneous vision and inertial sensors;
\cite{R20} fuses vision information with radar data and uses convolutional neural networks to detect key positions on the subway route for train positioning.
However, such methods relying on matching and locating key features along the route often perform poorly in tunnel scenes with simple patterns, repeated textures, and insufficient lighting. 
In this paper, we tackle this problem by utilizing the stable emergence of rail features.
We propose an aerial-view sleeper detection based railroad self-positioning method, 
incorporating the sensing data from on-board cameras and speed meters to provide accurate subway locations.
We show that with the proposed robust sleeper detection algorithm and geometric-constrained position estimation scheme, our subway self-positioning method can achieve stable and accurate underground self-localization at low cost and high speed.
A large-scale subway self-positioning dataset is collected to demonstrate the effectiveness of the proposed method, 
where we carry out extensive comparisons and analyses on quantitative experimental results.

Overall, the main contributions of this paper are three-fold:
\begin{itemize}
\item{we build a large-scale subway self-positioning dataset, the VSL Dataset (Vision-based Subway Localization Dataset) with subway front-view videos and corresponding ground-truth positions for comparison benchmarks;}
\item{we propose a robust and accurate sleeper detection algorithm based on aerial-view transformation and prior knowledge calibration, which can be easily adapted to other railway applications;}
\item{we design a visual-based, infrastructure-free subway self-localization method available for both open filed and underground scenarios.}
\end{itemize}
\section{Related work}
Different from ground vehicles, which can rely on the combination of IMUs, satellites, and odometers for precise positioning, subways running underground in tunnels lack GNSS signals. Consequently, they cannot obtain positions solely from IMUs and odometers due to the accumulation of errors.
As a result, subways usually have much more complicated positioning systems composed of track-side equipment (external devices) and advanced localization algorithms; while the problem is still not perfectly solved and has long been a research hot spot.
Based on whether external devices are needed, we can roughly categorize current research in the field of subway positioning into two types: external information-assisted positioning and self-positioning.

\subsection{External Information assisted Positioning}

In subways, externally provided ground truth positions are usually needed as correction either by hardware devices or by intelligent algorithms, in that traditional on-board sensors (e.g. wheel encoders, Doppler radars, IMUs) share the inherent characteristics of time-cumulative sensing errors. External information-assisted positioning methods rely on external guidance, signal sources, or other auxiliary information to determine the subject's locations.

One of the earliest technologies used for external information-assisted positioning in subway systems is the track circuit. This method utilizes two steel rails as conductors, forming an electrical circuit through connecting wires that link signal transmission and reception devices. Continuous tracking of the track circuit's occupancy status allows for the ongoing monitoring of the train's position on the rail route\cite{R53}. This rail-based positioning method\cite{R41, R42}, known for its resilience to environmental disturbances and relatively lower track circuit installation costs, was employed for an extended period as a coarse-grained subway positioning system. The limited positioning accuracy of track circuits has prompted researchers to increasingly explore responder-interrogator methods capable of providing higher precision.  For example, certain routes in the St Petersburg subway system are equipped with automatic train control systems supported by UHF RFID technology \cite{R24, R8, R10, R25}. This involves readers accessing the memory of RFID tags on tunnel walls wirelessly to obtain coordinate information (latitude and longitude) and control commands. To address sparse reduction and computational effort challenges during high-speed train movement, balise-based positioning algorithms are further improved by \cite{R26, R27} using a least squares support vector machine (LSSVM) model to reduce position errors. Another cost-effective approach for external information-assisted positioning is based on visible light communication (VLC) systems \cite{R43}. The real-time position calculation is achieved through the control of LED lights embedded in tunnel walls via an onboard transmitter, coupled with the analysis of light intensity information by an onboard receiver.

While the above methods can achieve relatively high localization accuracy, they exhibit a strong reliance on external infrastructure.  This dependence leads to significantly elevated installation and maintenance costs, and there is a potential for rapid loss of positioning accuracy in the face of external equipment failures.

\subsection{Self-positioning}
\label{sec:ref_self}

In contrast to external information assisted positioning techniques, self-positioning algorithms naturally remain immune to external interference, thanks to their independent localization procedures. Besides offering higher stability, self-positioning methods prove to be more cost-effective since they eliminate the need for pre-installed track-side devices, and maintenance doesn't necessitate a complete route overhaul. However, maintaining high precision with self-positioning methods is a greater challenge, as environmental sensing data from on-board sensors may lack distinctive features and become challenging to recognize, particularly within subway underground tunnels.

Researchers have tried various sensors and algorithms to make self-positioning methods as effective as device assisted ones. For example, \cite {R9} utilizes geomagnetic information and feature map matching based on passive magnetic intensity measurement; \cite{R31} fuses the sensing data of wave radar and IMUs to carry out a curvature estimation of subject positions, and \cite{R11} exploits the trajectory map constraints measured by a MEMS IMU and other sensors to predict navigation information. Over time, investigators have gradually found that LiDAR-based and vision-based sensors are the two most representative and effective sensing approaches for subway self-positioning methods.

The reason that LIDAR becomes a majority choice by high-performance method among the massive selectable on-board sensors, is that LIDAR has a strong ability to scan detailed environmental features and build digital maps. For example, \cite{R12} validates that a simple sliding window mapping based on LIDAR data can improve the subway positioning accuracy in tunnels; \cite{R32} collects subway distance information of different civil construction surfaces by multiple sets of LIDAR devices, and shows the capability of feature matching localization based on various LIDAR sensors. \cite{R48} reached sub-meter-level positioning accuracy solely relying on LIDAR point cloud data, further validating the efficacy of LIDAR signals. By integrating LIDAR with inertial navigation systems and track control networks, \cite{R33} can even achieve millimeter-level measurement and mapping of tunnel railways at speeds ranging from 0.5 to 1.2 m/s.While LiDAR is renowned for its high precision, its high sensitivity to dynamic objects and weather conditions, coupled with elevated costs, limit its extensive usage primarily to the fields of track maintenance and mapping.

The other highlighted sensor for subway positioning is the camera, which can provide low-cost and richly informative visual sensing data. Back in the days when cameras could only provide low-resolution images, they were already being used in train positioning systems to rectify train poses and select track routes. For example, \cite{R36} extracts the rail space from images through dynamic planning for train pose estimation, and \cite{R37} autonomously selects train track routes by detecting and tracking angle changes in images. Nowadays, with the development in both camera devices and intelligent algorithms, visual information can achieve more elaborate estimations\cite{R54}, like the SLAM (Simultaneous Localization And Mapping) methods in the domain of autonomous driving \cite{R38}. \cite{R19} demonstrates that the high-performing visual SLAM algorithms are also effective on trains (in well-lit outdoor environments) by evaluating several advanced algorithms in a railroad dataset. However, as shown in \cite{R44}, SLAM methods exhibit severe performance degradation when conducted on rail transit scenarios with repetitive textures and varying lighting conditions. To overcome the challenges of texture-constrained subway localization, \cite{R20} employed key position feature extraction through the integration of deep neural networks. They fused visual information with millimeter-wave radar data for tunnel positioning, achieving a significant reduction in the maximum localization error to 4.7 meters on a total route length of 16 kilometers.

In this paper,
we tackle the problem of vision-based subway self-positioning, 
based on an aerial view rail sleeper detection network and real-time localization estimation.
\begin{figure*}[!t]
\centering
\includegraphics[width=7in]{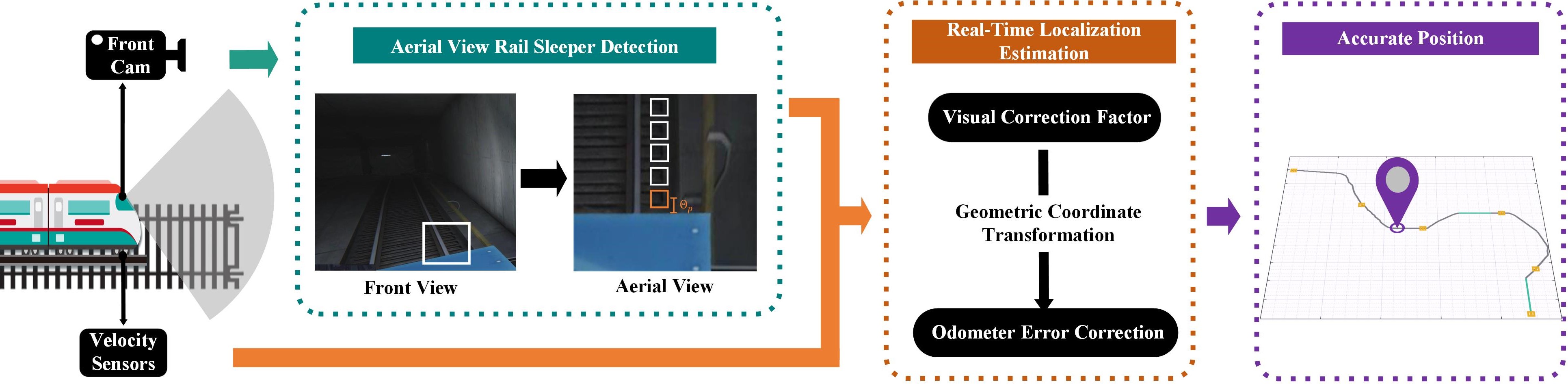}
\caption{Framework of the proposed subway self-positioning system. 
The entire framework takes in real-time inputs of the forward camera and speed sensor. 
Aerial view rail sleeper detection module performs visual transformation on the input image and detects the positions of sleepers on the railway track. 
Real-time localization estimation module utilizes the visual correction factor to periodically correct the accumulating errors in the mileage. 
Ultimately, the framework outputs the precise position of the subway train on its route.}
\label{fig:fig_1}
\end{figure*}

\section{Methodology}
\label{sec:method}
In this section, we introduce the framework of the proposed real-time, metro-self-localization system, 
which comprises two principal components as depicted in Fig. \ref{fig:fig_1},
i.e. aerial view rail sleeper detection module 
and real-time localization estimation module.
To be specific, 
the aerial view rail sleeper detection module firstly transforms images captured by a front camera of the subway train to a unified aerial view,
and then feed the transformed and cropped image of single-sided rail sleepers into a detection network for rail sleeper position detection; 
with the accurately detected sleeper positions, the geometric inference based position estimation module utilizes geometric relationships for pixel-to-world coordinate transformation 
to deduce the exact system position referring to the nearest rail sleeper ($\theta$ meters from the front of the subway).
Note that $\theta$ is subsequently employed for real-time correction of accumulated errors, 
allowing the system to maintain a long-lasting high self-positioning precision.

\subsection{Aerial View Rail Sleeper Detection Module}
\label{sec:method_IPM}
One distinctive feature of the subway circumstances is that all trains are 
constrained within the railroad tracks. 
Therefore, we propose that the rail sleeper, 
as a basic and frequent component of railroad tracks,
can make effective location markers for subway self-positioning.
In the aerial view rail sleeper detection module, we explore the position estimation of subway trains from sleeper information,
where we carry out accurate sleeper detection based on deep network,
and equip the module with subway-specific prior knowledge to boost detection precision and stabilization.
The overall module consists of two main steps.

\noindent {\bf{Step 1. Fast View Transformation}}
Before sleeper detection, we first transform the front view images (captured from front cameras in any attitude) to a unified aerial view.
The reasons we choose aerial view images as system inputs are that prior restraint of rail tracks are easier to apply in aerial ways, 
and that aerial view images can provide more practical low-dimensional representation as demonstrated in the field of autonomous driving \cite{R39}. 
Mainstream image view transformation methods often incorporate inverse perspective mapping or affine transformation to transform input images from one view point to another.
Compared with affine transformation which usually finds the transformation matrix via forcibly aligning source and target quadrangles,
inverse perspective mapping appears to be more consistent with the imaging principle as it estimates a three-dimensional transformation that first maps the source image pixels back to the 3D world coordinates and then calculates their target imaging positions accordingly. 
Therefore,
we accomplish the view transformation of subway images with a fast inverse perspective mapping (IPM) method.

\begin{figure}[!t]
\centering
\includegraphics[width=3.5in,height=3.25in]{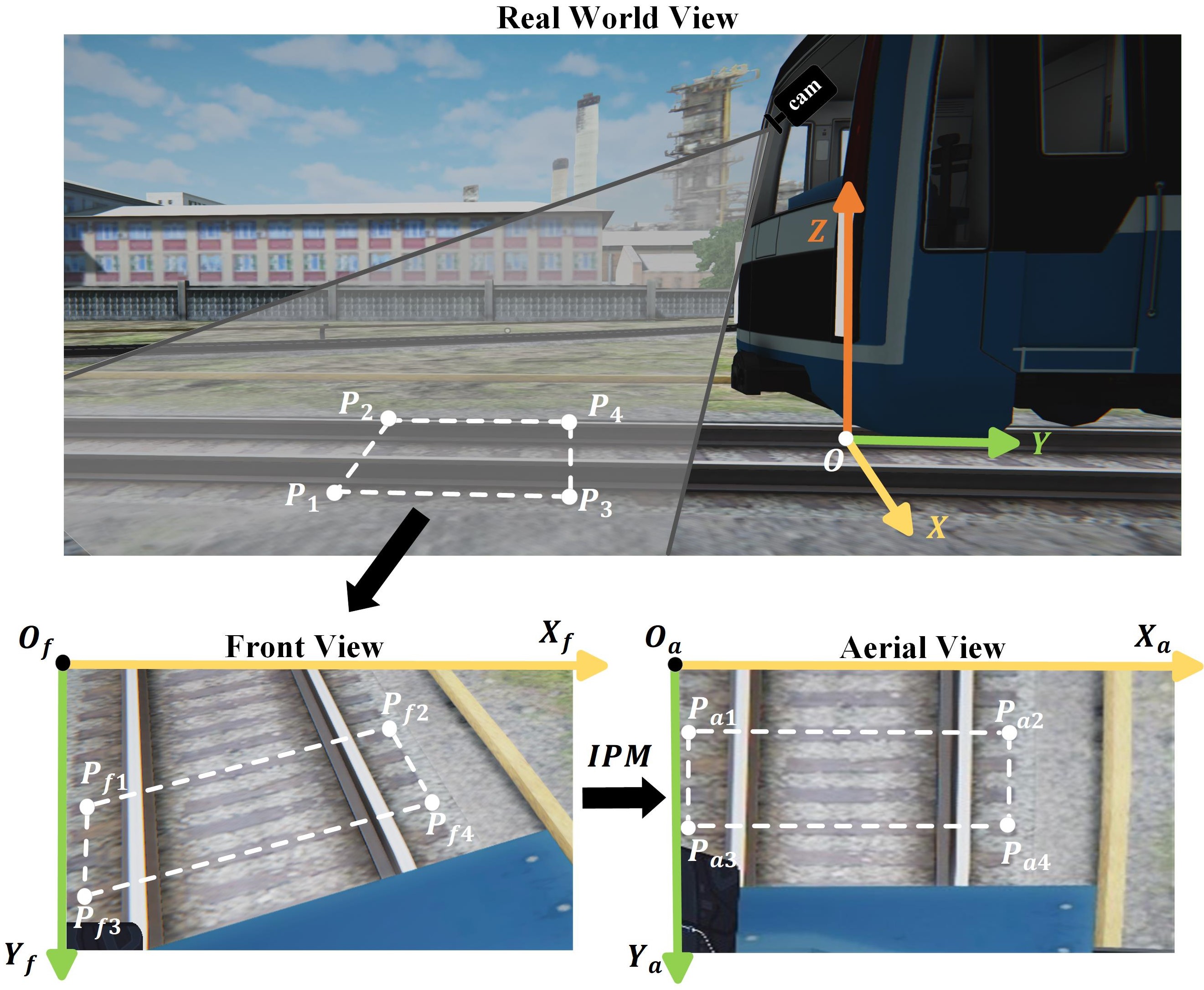}
\caption{Image perspective conversion process (IPM). Illustration of the mapping relationships and the transformation process among real-world coordinate system ($O-XYZ$), front view image plane ($X_f-O_f-Y_f$), and aerial view image plane ($X_a-O_a-Y_a$).}
\label{fig:fig_2}
\end{figure}

Although the camera parameters are unknown in our subway videos,
we propose that as subway tunnels have very similar depth distributions,
a simple and fast four-point-conversion IPM can achieve the required mapping from front view to aerial view.
As shown in Fig. \ref{fig:fig_2}, 
four particular points are picked and recorded from the general front view image for estimation of the transformation metric,
which is located in the corners of the bounding box covering the second to the fifth sleeper.
The pixel coordinates in front view image of these four points $P_{f_{1}}, P_{f_{2}}, P_{f_{3}}, P_{f_{4}}$ are $(x_{f_{1}},y_{f_{1}}), (x_{f_{2}},y_{f_{2}}), (x_{f_{3}},y_{f_{3}})$ and $(x_{f_{4}},y_{f_{4}})$, respectively.
Assume that the point in the world coordinate system corresponding to $P_{f_{i}}$ is $P_{i}$ ($X_{i}, Y_{i}, Z_{i}$), and the point in the aerial view is $P_{a_{i}}$ ($x_{a_{i}},y_{a_{i}}$), $i=1,2,3,4$;
the real-world coordinate system has X and Y axes that are parallel and vertical to the vehicle's front platform,
original point $O (0,0,0)$ that is located at the ground projection of the front platform center;
and the Z axis set according to the right-hand rule (As shown in Fig. \ref{fig:fig_2}). 
As $P_{i}$ is placed on the $X-O-Y$ plane, we have $Z_{i} = 0$. 
According to the principle of camera perspective transformation \cite{R57}, we can establish the following relationship between image-plane coordinates and world coordinates:
\begin{align}
\begin{bmatrix}
x_{f_{i}} \\
y_{f_{i}} \\
1\\
\end{bmatrix}
=K_{f}{\frac{1}{Z_{f}}}{P_{{w}\rightarrow{f}}}
\begin{bmatrix}
X_{i} \\
Y_{i} \\
Z_{i} \\
1\\
\end{bmatrix}
\label{eq:1}
\end{align}
\begin{align}
\begin{bmatrix}
x_{a_{i}} \\
y_{a_{i}} \\
1\\
\end{bmatrix}
=K_{a}{\frac{1}{Z_{a}}}{P_{{w}\rightarrow{a}}}
\begin{bmatrix}
X_{i} \\
Y_{i} \\
Z_{i} \\
1\\
\end{bmatrix}
\label{eq:2}
\end{align}

\noindent where $_a$, $_f$ and $_w$ denote the aerial view image plane, front view image plane, and real world 3D coordinate system respectively;
$K_{a}$ and $K_{f}$ are the $3\times3$ camera intrinsic parameter matrixes ($K_{a}=K_{f}$ in this case);
$Z_{a}$ and $Z_{f}$ are the constants for object distances;
and $P$ represent the $3\times4$ camera extrinsic parameter matrix 
between real world and aerial view image plane ${P_{{w}\rightarrow{a}}}$,
and between real world and front view image plane ${P_{{w}\rightarrow{f}}}$.
From equations \ref{eq:1} and \ref{eq:2}, we have:
\begin{align}
Z_aP_{a\rightarrow w}^{4\times3}K^{-1}\left[\begin{matrix}x_{a_{i}}\\y_{a_{i}}\\1\\\end{matrix}\right]
=
Z_fP_{f\rightarrow w}^{4\times3}K^{-1}\left[\begin{matrix}x_{f_{i}}\\y_{f_{i}}\\1\\\end{matrix}\right]
\end{align}

\begin{align}
\left[\begin{matrix}x_{a_{i}}\\y_{a_{i}}\\1\\\end{matrix}\right]
=
\frac{Z_f}{Z_a}P_{a}^{3\times4}P_{w}^{4\times3}\left[\begin{matrix}x_{f_{i}}\\y_{f_{i}}\\1\\\end{matrix}\right]\ =\ H^{3\times3}\left[\begin{matrix}x_{f_{i}}\\x_{f_{i}}\\1\\\end{matrix}\right]
\end{align}

\noindent where $H$ is a fixed $3\times3$ parameter matrix.
We obtain the estimated values in $H$ 
by solving the system of equations using manually labeled pairs of non-collinear points (as shown in Fig. \ref{fig:fig_2}).
As the front camera on subway trains usually has fixed installation locations and angels,
the estimated transformation matrix $H$ can be applied to all images without adjustment.

\noindent {\bf{Step 2. Rail Sleeper Detection}}
Through a thorough survey, we identify YOLOv8 \cite{R50} as one of the most well-established and advanced target detection networks. 
YOLOv8 offers anchor-free detection model architectures of various sizes and complexities (e.g. n, s, m, l) tailored to diverse usage scenarios;
among them, the YOLOv8n network stands out as the smallest and fastest model which maintains excellent detection accuracy.
Therefore, we incorporate the YOLOv8n structure as rail sleeper detector in the proposed positioning framework.
In Section \ref{sec:exp_sleeper_detection},
we show that this detection net is particularly well-suited for scenarios like subway driving where real-time requirements are critical.
As the proposed framework relies on the nearest rail sleeper and its distance to the train front for position rectification and the orbitals have a symmetrical structure, we use the single-sided aerial-view images as inputs for efficient sleeper detection.

During the training of the rail sleeper detector, we manually annotated the positions of all sleepers in the aerial-view images, i.e. each sleeper is enclosed by a fixed-size square bounding box ($s \times s$ centered at $S_{i}$ like shown in Fig. \ref{fig:fig_3}). 
The rail sleeper detector is fine-tuned with the annotated samples from YOLOv8n pre-trained on the VOC dataset \cite{R58}. 

In the inference process, the aerial view sleeper detection module transforms and crops front view images to single-sided aerial view images, and then sequentially feeds them to the sleeper detector to extract visual references.
These references provide fundamental positioning information for error correction in the following localization estimation module. 
We show quantitative evaluation results of the sleeper detector in Section \ref{sec:exp_sleeper_detection},
verifying the effectiveness and high accuracy of the sleeper detection module.

\subsection{Real-Time Localization Estimation Module}
\label{sec:module_localization}
\begin{figure}[!t]
\centering
\includegraphics[width=3.5in,height=3.8in]{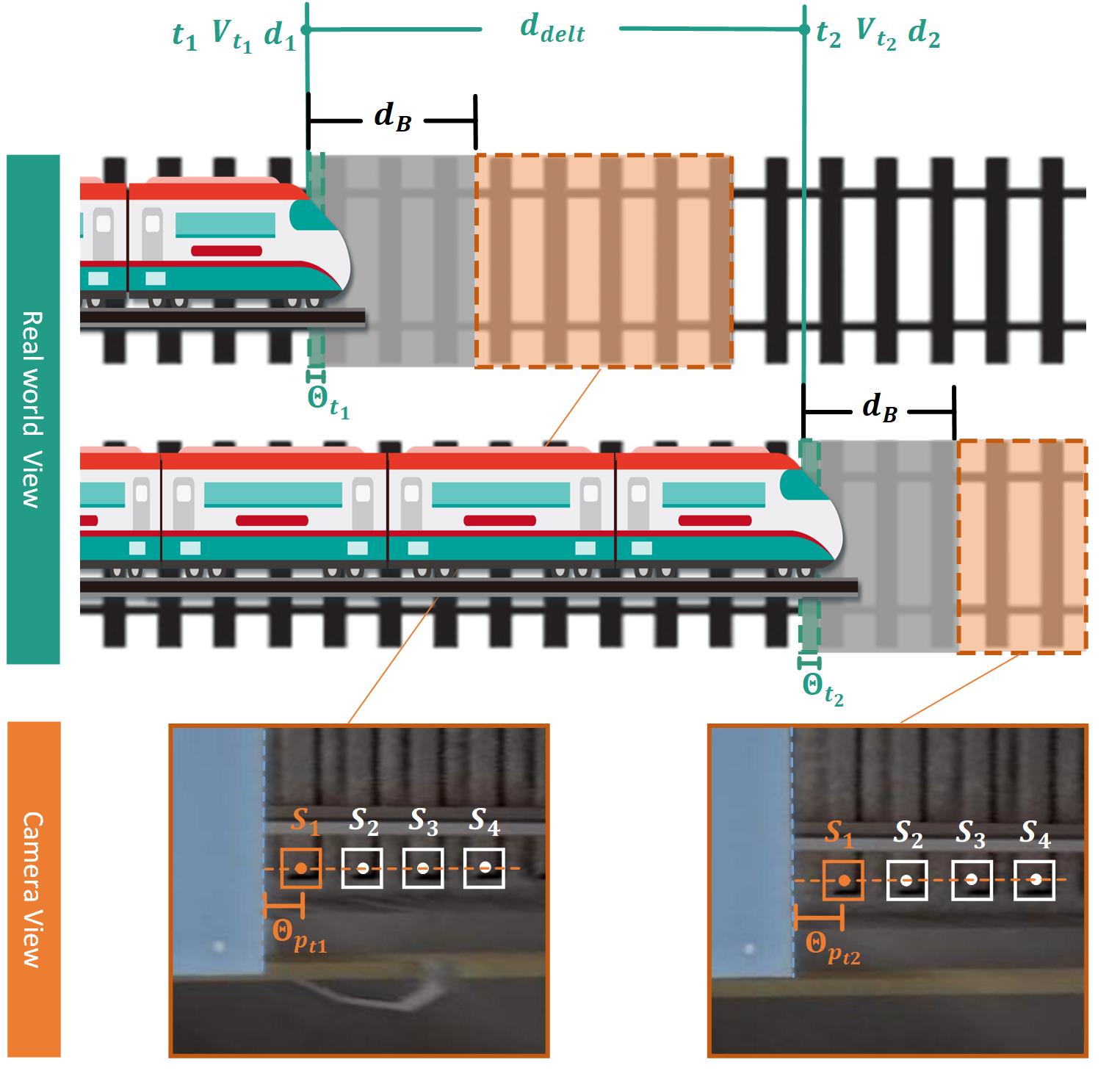}
\caption{Illustration of the real-time localization estimation process. 
In adjacent moments $t_{1}$ and $t_{2}$,
sleepers captured by the front camera are detected in the aerial view (Camera View), 
where the nearest sleeper provides a visual correction factor to estimate the exact advance distance for a subway train (Real World View).
}
\label{fig:fig_3}
\end{figure}
The real-time localization estimation module corrects the accumulated sensor error by calculating the distance from the closest rail sleeper to the front of the subway, thereby estimating the accurate position of the subway relative to the entire route.
The key process is illustrated in Fig. \ref{fig:fig_3}, which shows the position information of the subway in both real-world and camera view coordinate systems during consecutive time points $t_{1}$ and $t_{2}$. 
Assuming that the subway positions at time $t_{1}$ and $t_2$ are $d_{1}$ and $d_2$, respectively;
we calculate the visual correction factors $\Theta_{t_{1}}$ and $\Theta_{t_{2}}$ based on the detection results of aerial view rail sleeper detection module. 
This rectification process effectively corrects the cumulative sensor error, allowing us to estimate $d_{\text{delta}}$ (distance between $d_1$ and $d_2$).
Consequently, at time $t_{2}$, the subway's position $d_{2}$ can be inferred.

The process of obtaining visual correction factors is as follows.
In the real scenario (as shown in Fig. \ref{fig:fig_3}), 
the front camera has a blind spot of a fixed size (the gray area).
Assuming its field of view starts $d_{B}$ meters in front of the train,
we use the transformed aerial view over the equipment platform (the blue rectangular area) for sleeper detection.
Based on the detected sleepers (e.g. $S_{1}, S_{2}, S_{3}, S_{4}$),
we choose the one nearest to the train front ($S_{1}$) for distance measurement to provide visual rectifications.

Set the distance in pixels between the sleeper $S_{1}$ and the device platform in the image to be $\Theta_{p}$, and its corresponding real-world coordinate distance denoted to be $\Theta_{r}$;
as the subway route is perpendicular to the train front,
distances $\Theta_{p}$ and $\Theta_{r}$ can be directly calculated by the coordinate difference on a single axis, i.e. $\Delta y$ and $\Delta Y$ in the coordinate system $X_a-O_a-Y_a$ and $O-XYZ$ in Fig. \ref{fig:fig_2}, respectively.
According to Equation \ref{eq:2}, 
there is:
\begin{align}
\begin{bmatrix}
0 \\
{\Delta y}\\
0\\
\end{bmatrix}
=K_{a}{\frac{1}{Z_{a}}}{P_{{w}\rightarrow{a}}}
\begin{bmatrix}
0 \\
{\Delta Y}\\
0\\
0\\
\end{bmatrix}
=
[a_{ij}]^{3\times4}
\begin{bmatrix}
0 \\
{\Delta Y}\\
0\\
0\\
\end{bmatrix}
\end{align}
\begin{align}
{\Delta y} = (a_{21}+a_{22}+a_{23}+a_{24})\cdot{\Delta Y} = r\cdot{\Delta y}
\label{eq:6}
\end{align}

\noindent
where \{$[a_{ij}]^{3\times4} | i=1,2,3; j=1,2,3,4$\} denotes a 3 by 4 matrix corresponds to $K_{a}{\frac{1}{Z_{a}}}{P_{{w}\rightarrow{a}}}$;
${\Delta y}$ and ${\Delta Y}$ represent the distances from the first seen sleeper to the train device platform in aerial view image and real-world respectively.

From Equation \ref{eq:6},
we can obtain the relationship between $\Theta_{r}$ and $\Theta_{p}$:
\begin{align}
\Theta_r=\frac{1}{r}\cdot\Theta_p
\end{align}

\noindent
where $\frac{1}{r}$ represents a fixed constant associated with the camera intrinsic and extrinsic matrices $K{\frac{1}{Z_{a}}}{P_{{w}\rightarrow{a}}}$. 
Based on the above mathematical reasoning,
$\Theta_{p}$ is proportional to $\Theta_{r}$ at a fixed rate.
In the proposed method, $r$ is estimated the same way as transformation matrix $H$ as described in the previous step.
At this point, 
we are able to compute the pixel distance perpendicular to the vehicle front front detection results,
and then estimate the corresponding real-world distance $\Theta_{r}$,
as well as the sleeper spacing $\tau$ and the blind area $d_{B}$, and etc.

The estimated $\Theta_{r}$ is used to compute the visual correction factor.
When $\Theta_{r} < \tau$, 
the detection model successfully obtains the nearest sleeper position, 
the visual correction factor $\Theta$ is computed with: 
\begin{align}
\Theta = \Theta_{r} + d_{B} - floor(\frac{\Theta_{r}+d_{B}}{\tau}) \cdot \tau
\end{align}
\noindent where $floor(.)$ denotes the downward rounding function. 

As there might be a small probability of failed detections (causing $\Theta_{real}>=\tau$), 
we also define the following equation to deal with such cases:
\begin{align}
\Theta=\xi\cdot\tau 
\end{align}
where $\xi\in(0,1)$, is a constant covariate set according to empirical values.

Given the velocities $V_{t_{1}}$ and $V_{t_{2}}$ at instants $t_{1}$ and $t_{2}$, the subway's movement $d_{delt}$ during the time $T = t_{2} - t_{1}$ can be expressed as two components, i.e. $L$ complete sleeper spacings plus the visual correction distance $\gamma$:
\begin{align}
d_{delt}=L\cdot \tau+ \gamma
\label{eq:10}
\end{align}
\noindent
where integer $L$ represents the number of passed sleepers during $T$. 
When $t_1$ and $t_2$ are close enough to surpass the error accumulation speed of speed sensors,
$L$ can be accurately computed by:
\begin{align}
L=floor(\frac{d_{V_T}}{\tau})
\end{align}
\noindent 
where $d_{V_T}$ is the distance between $t_1$ and $t_2$ calculated by the speed sensor: $d_{V_T}=0.5\cdot(V_{t_{1}}+V_{t_{2}})\cdot T$.

With the acquired visual correction factors $\Theta_{t_{1}}$ and $\Theta_{t_{2}}$ (as shown in Fig. \ref{fig:fig_3}), 
the remainder part $\gamma$ in Equation \ref{eq:10} is discussed in two distinct scenarios:
\begin{align}
\gamma = \begin{cases}
\Theta_{t_{2}}-\Theta_{t_{1}} & \Theta_{t_{2}}>=\Theta_{t_{1}}  \\
\Theta_{t_{2}}-\Theta_{t_{1}}+\tau & \Theta_{t_{2}}<\Theta_{t_{1}} \\
\end{cases}
\end{align}

As a result, 
the proposed method is able to continuously rectify mileage errors through the calculated value of $d_{delt}$ based on the visual correction factor,
and achieves substantial enhancement of real-time positioning accuracy.
%

%
\section{EXPERIMENTS}
\subsection{Subway Localization Dataset}
\label{sec:dataset}
\begin{figure}[!t]
\centering
\includegraphics[width=3.2in,height=4in]{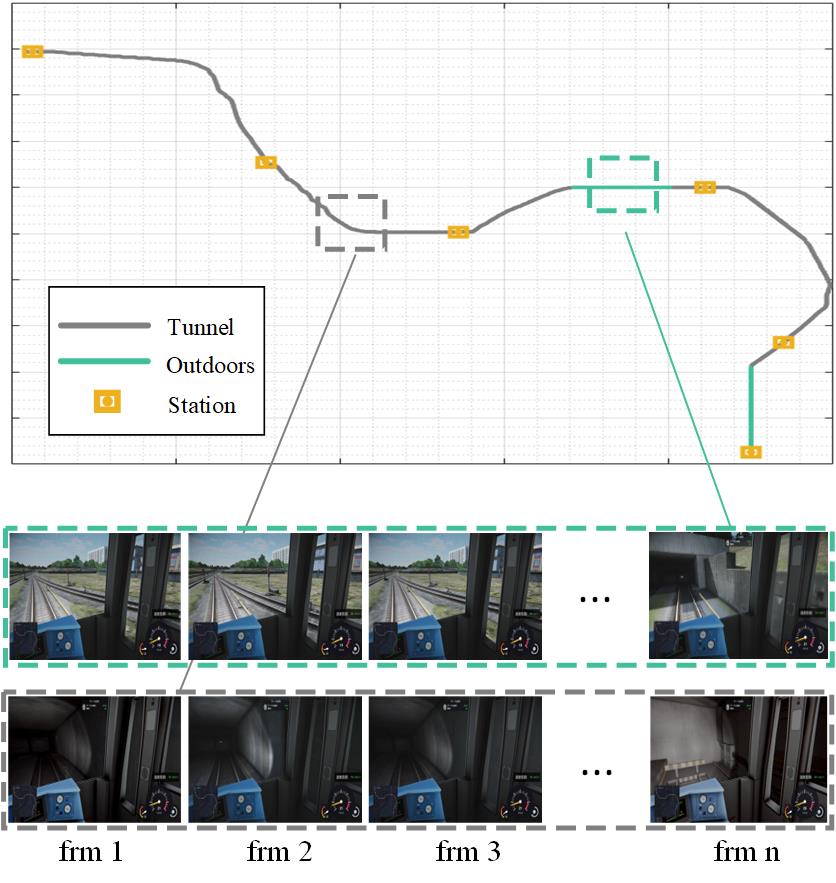}
\caption{Illustration of the subway route and video frames (from front camera).
The total route is about 6.9 km with color gray for the tunnels and color green for non-tunnel environments. 
Two typical example videos for the train running in and outside of subway tunnels are shown below, colored in correspondence with the route. 
}
\label{fig:fig_4}
\end{figure}

Due to the challenges in obtaining real-world subway data,
we establish a 
Vision-based Subway Localization (VSL) Dataset using the "Subway Simulator 3D" game engine.
The total route is about 6.9 kilometers in length,
where the majority environment is inside the tunnel (around 6.3 kilometers).
As illustrated in Fig. \ref{fig:fig_4},
the entire route comprises six stations,
whose inter-station distances are about 1.1km, 1.3km, 1.6km, 1.3km, and 1.6km, respectively.
We collect the driving video along the whole route at 15 fps and manually annotate the ground-truth localizations for each video frame (18392 frames in total).
Illustrations of the subway route and video frames in and out of tunnels are shown in Fig. \ref{fig:fig_4}.
Figure \ref{fig:fig_4} showcases partial video frames corresponding to scenes within the tunnel and outdoors.

\begin{figure}[htbp]
    \centering
    \subfigure[Different Lighting Conditions]{
        \includegraphics[width=0.45\textwidth]{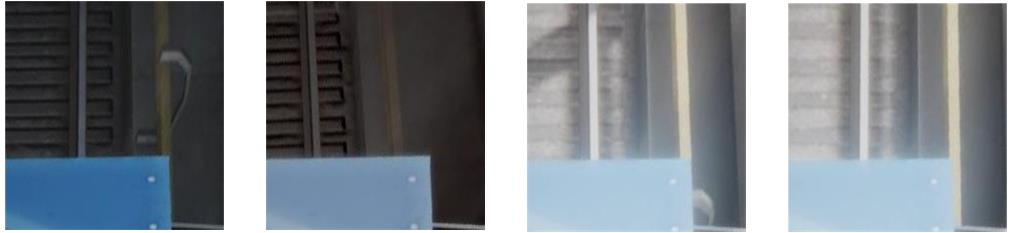}
    }
    \subfigure[Different Road Conditions]{
        \includegraphics[width=0.45\textwidth]{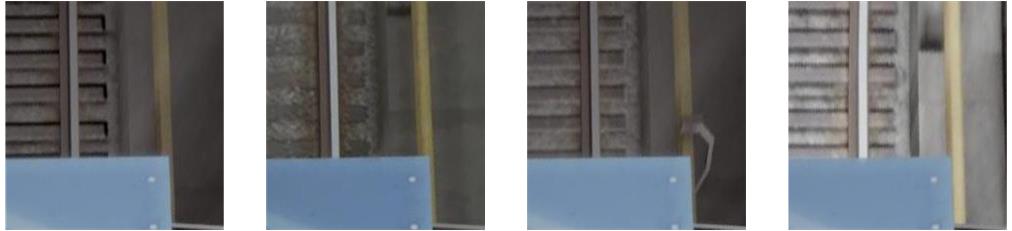}
    }
    \caption{Examples for aerial view images of rail sleepers. %
    (a) and (b) show cases of different lighting and road conditions respectively.}
    \label{fig:fig_5}
\end{figure}

To train the sleeper detector, 
we also build an aerial view rail sleeper dataset containing 849 images with high-quality sleeper annotations.
Each image is generated by aerial view transformation (described in Section \ref{sec:method_IPM}) from the original subway videos, 
cropped with the same size of $256 \times 256$ along the equipment platform side. 
As shown in Fig \ref{fig:fig_5},
this sleeper dataset covers a large variety of lighting conditions and road patterns.
Nevertheless, these images share a uniform aerial view standard, 
demonstrating the effectiveness of our fast view transformation algorithm (Section \ref{sec:method_IPM}).
We annotate each sleeper position in the images,
and use 80\% and 20\% for training and evaluation of the sleeper detector respectively.

We employ Precision (P), Recall (R), and F1-score (F1) to evaluate the performance of the rail sleeper module.
P and R are two widely used metrics in the field of object detection,
where P represents the proportion of correctly predicted positive samples in all predicted samples;
and R represents the proportion of ground truth samples that have been correctly detected.
F1 is a balance score between P and R calculated with $F1=(2 \times \text{P} \times \text{R})/(\text{P} + \text{R})$,
with a higher score (closer to 1) indicating better algorithm performance.

For assessment of the positioning system, 
we use the Maximum Error (ME) and Mean Percentage Error (MPE) to respectively evaluate the extremal and mean levels of measurements:
\begin{align}
ME &=\max(|P_m^i-P_r^i|) \\
MAE &=\ \frac{1}{N}\sum_{i=1}^{N}\frac{{|P}_m^i-P_r^i|}{P_r^i}
\end{align}
\noindent
where N represents the total number of measurement points, $P_m^i$ denotes the measured mileage value of the positioning system at the $i-th$ point, and $P_r^i$ represents the true mileage value at the $i-th$ point.

Both datasets will be made available to the public to facilitate further studies in subway positioning.

\subsection{Aerial View Rail Sleeper Detection}
\label{sec:exp_sleeper_detection}
In our tunnel subway self-positioning scheme, sleeper detection is a crucial step. 
real time and accurate sleeper detection is of decisive significance. 
In this part, we conduct quantitative analysis on the sleeper detection task with various detectors, validating the effectiveness of the chosen YOLOv8n-based detector.

\begin{figure}[!t]
\centering
\includegraphics[width=3.3in,height=2.85in]{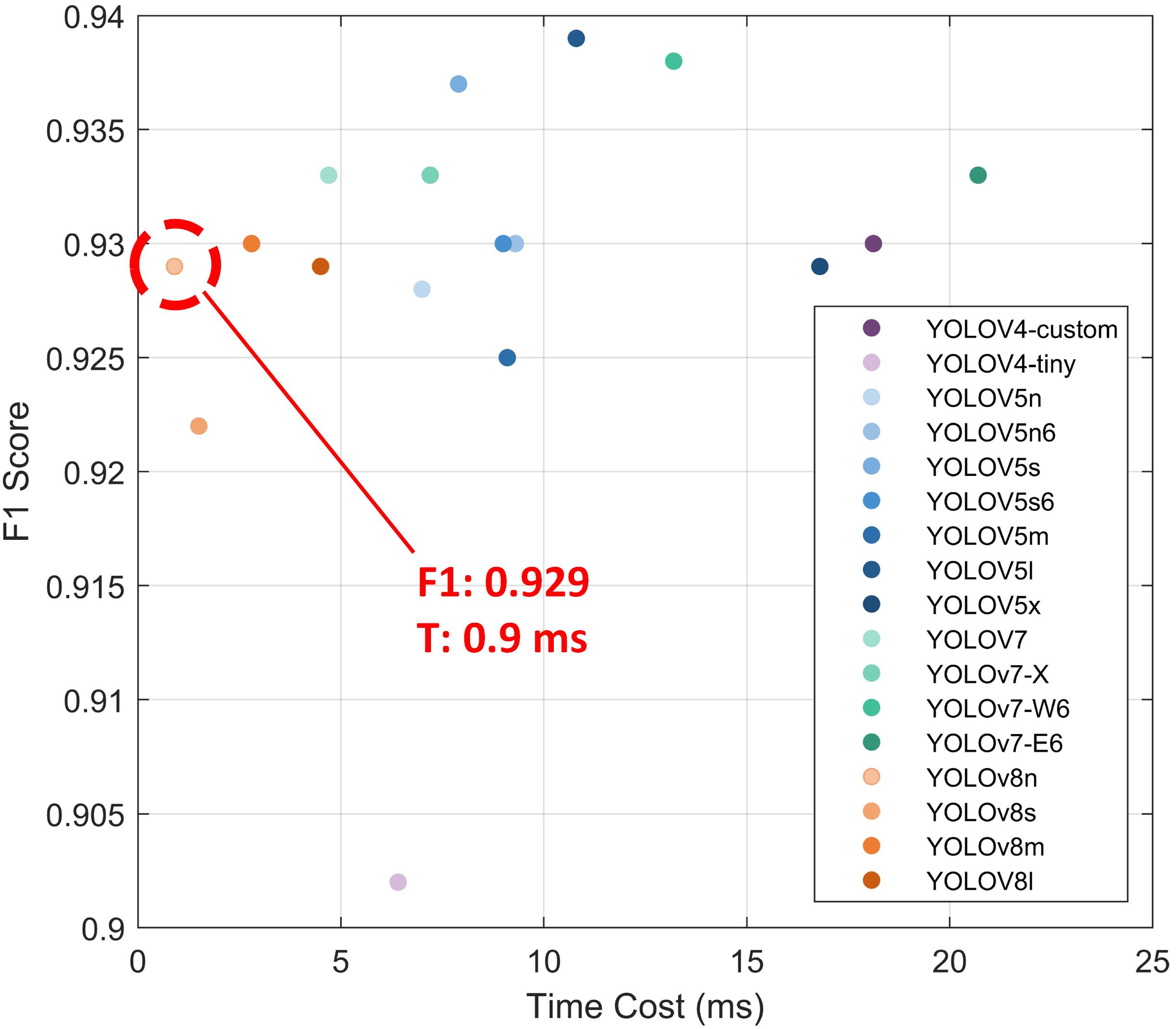}
\caption{Performance of different detection algorithms on aerial view of rail sleeper datasets (F1 score-Time Cost). 
The selected detection network in our aerial view sleeper detection module is highlighted with the red circle.}
\label{fig:fig_6}
\end{figure}
\begin{table}[h]
 \renewcommand{\arraystretch}{1.6}
 \caption{Sleeper Detection Performance\label{tab:table1}}  
    \centering
    \begin{tabular}{llccc}
        \toprule
        Method & P & R & F1 & Tims Cost (ms)  \\
        \midrule
        YOLOv4-custom  & 0.924 & 0.935 & 0.930 & 18.12 \\
		YOLOv4-tiny  & 0.912 & 0.893 & 0.902 & 6.41 \\
        YOLOv5n  & 0.931 & 0.925 & 0.928 & 7.00 \\
		YOLOv5n6  & 0.917 & 0.943 & 0.930 & 9.30 \\
		YOLOv5s  & 0.923 & 0.952 & 0.937 & 7.90 \\
		YOLOv5s6  & 0.917 & 0.943 & 0.930 & 9.00 \\
		YOLOv5m  & 0.920 & 0.931 & 0.925 & 9.10 \\
		\textbf{YOLOv5l}  & \textbf{0.927} & \textbf{0.951} & \textbf{0.939} & \textbf{10.80} \\
		YOLOv5x  & 0.924 & 0.935 & 0.929 & 16.80 \\
        YOLOv7  &0.907 &  0.960 & 0.933 & 4.70 \\
		YOLOv7-X  &0.944 &  0.923 & 0.933 & 7.2 \\
		YOLOv7-W6  &0.935 &  0.941 & 0.938 & 13.2 \\
		YOLOv7-E6  &0.954 &  0.913 & 0.933 & 20.7 \\
        \textcolor{blue}{YOLOV8n}  & \textcolor{blue}{0.930} & \textcolor{blue}{0.928} & \textcolor{blue}{0.929} & \textcolor{blue}{0.90} \\
		YOLOv8s  & 0.950 &0.896 & 0.922 & 1.50 \\
		YOLOv8m  & 0.937 &0.923 & 0.930 & 2.80 \\
		YOLOv8l  & 0.935 &0.923 & 0.929 & 4.50 \\		
        \bottomrule
    \end{tabular}
\end{table}
To assess the effectiveness of the algorithm, we train and test various YOLO networks (YOLOv4 \cite{R40}, YOLOv5 \cite{R45}, and YOLOv7 \cite{R49}) using the same dataset,
the overall performances on the test set are shown in Table \ref{tab:table1},
where we comprehensively compare the P, R, and F1 scores as well as the time costs (detection time per image) of different models.
For a more intuitive depiction,
we also draw the F1-Time Cost curve in Fig. \ref{fig:fig_6}.
From Fig. \ref{fig:fig_6},
we can clearly see that in comparison to other algorithms, the YOLOv8n-based algorithm exhibits optimal real-time performance (0.9 ms per frame) while maintaining high precision (less than 1\% lower to the best detection model but 12.3 ms faster).
This is attributed to the lightweight and low-parameter characteristics of YOLOv8n, e.g. compared with YOLOv5l, YOLOv8 utilizes a more enriched C2f structure with more gradient flow to ensure lightweightness; its channel numbers are adjusted according to the model scale, significantly improving the model performance and enhancing the real-time efficiency.
Another reason is that the straightforward characteristics of the aerial view sleeper images allow appropriate fitting by a compact and low-complexity network (even the worst-performing method in Table \ref{tab:table1} has F1 over 90\% in this specific task).
Therefore, 
it is adequate and reasonable that we ultimately choose the YOLOv8n-based detection network for high-precision and real-time sleeper detection modules.

\begin{figure*}[t]
	\centering
	\vspace{-0.15in}
	\begin{minipage}{1\linewidth}	%
		\subfigure[Ground Truth]{
			
			\includegraphics[width=3.5in,height=2in]{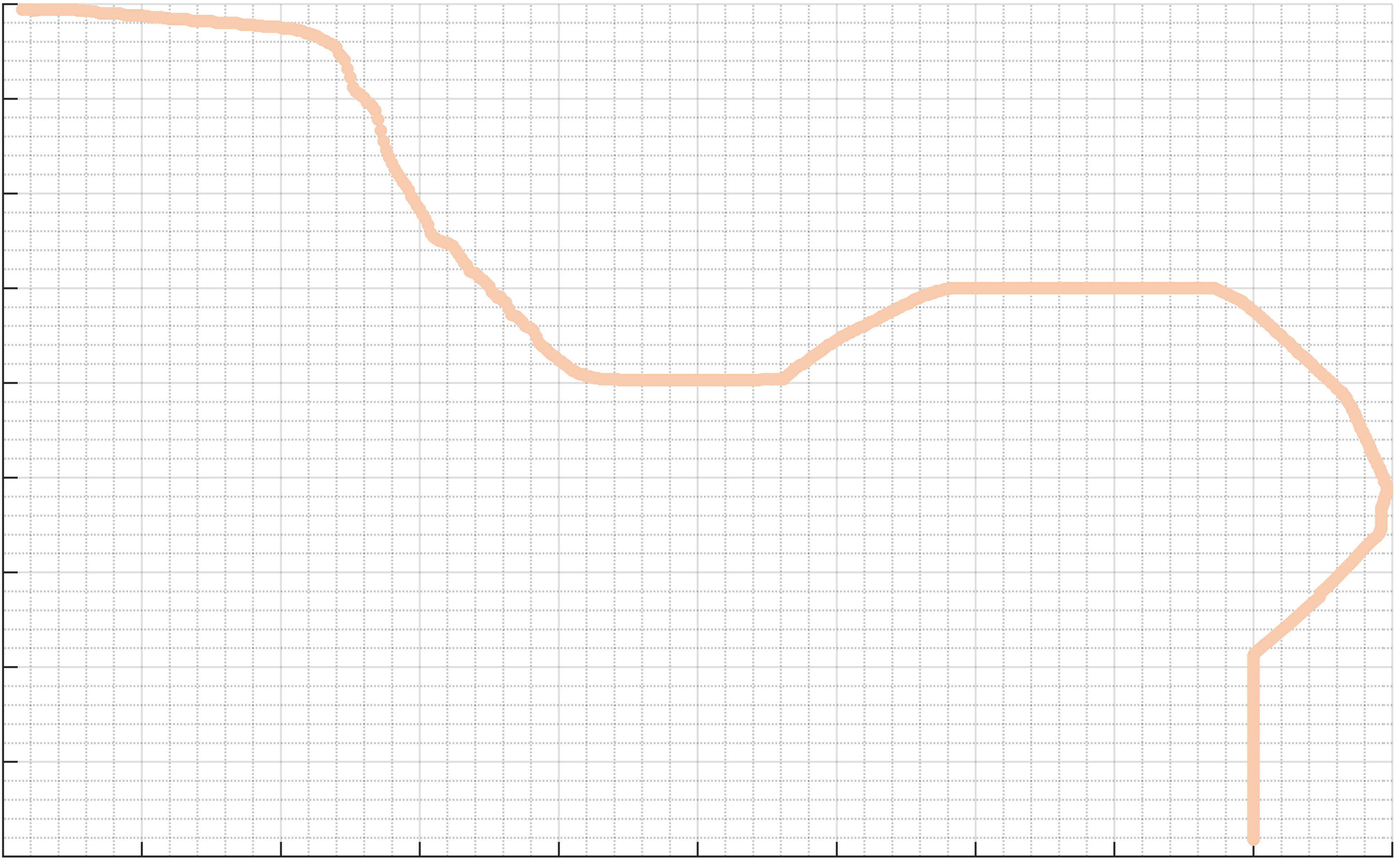}	
		}\noindent
		\subfigure[ORB-SLAM3]{
			\includegraphics[width=3.5in,height=2in]{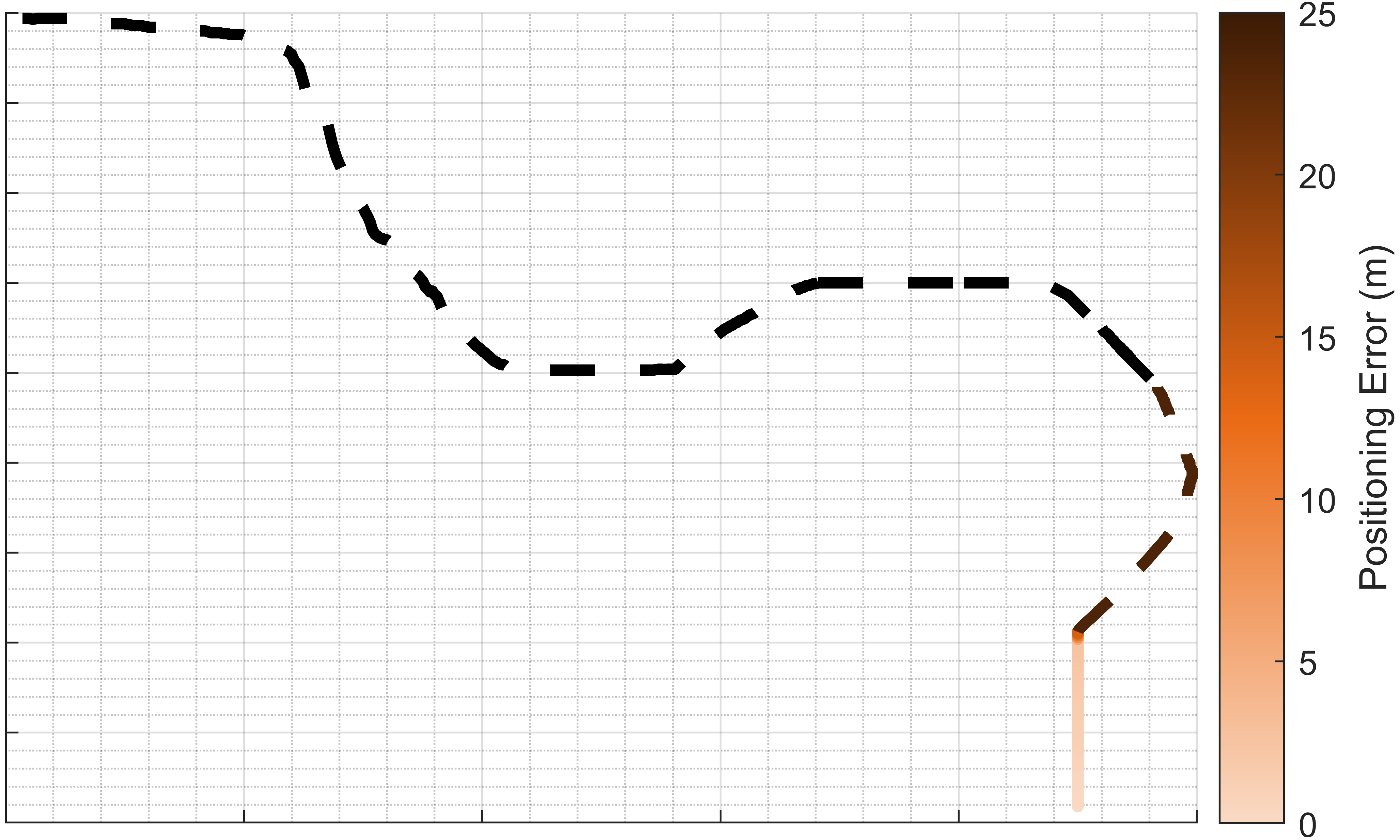}	
		}
	\end{minipage}
	\vskip -0.3cm %
	\begin{minipage}{1\linewidth }
		\subfigure[Direct Integral]{
			\includegraphics[width=3.5in,height=2in]{{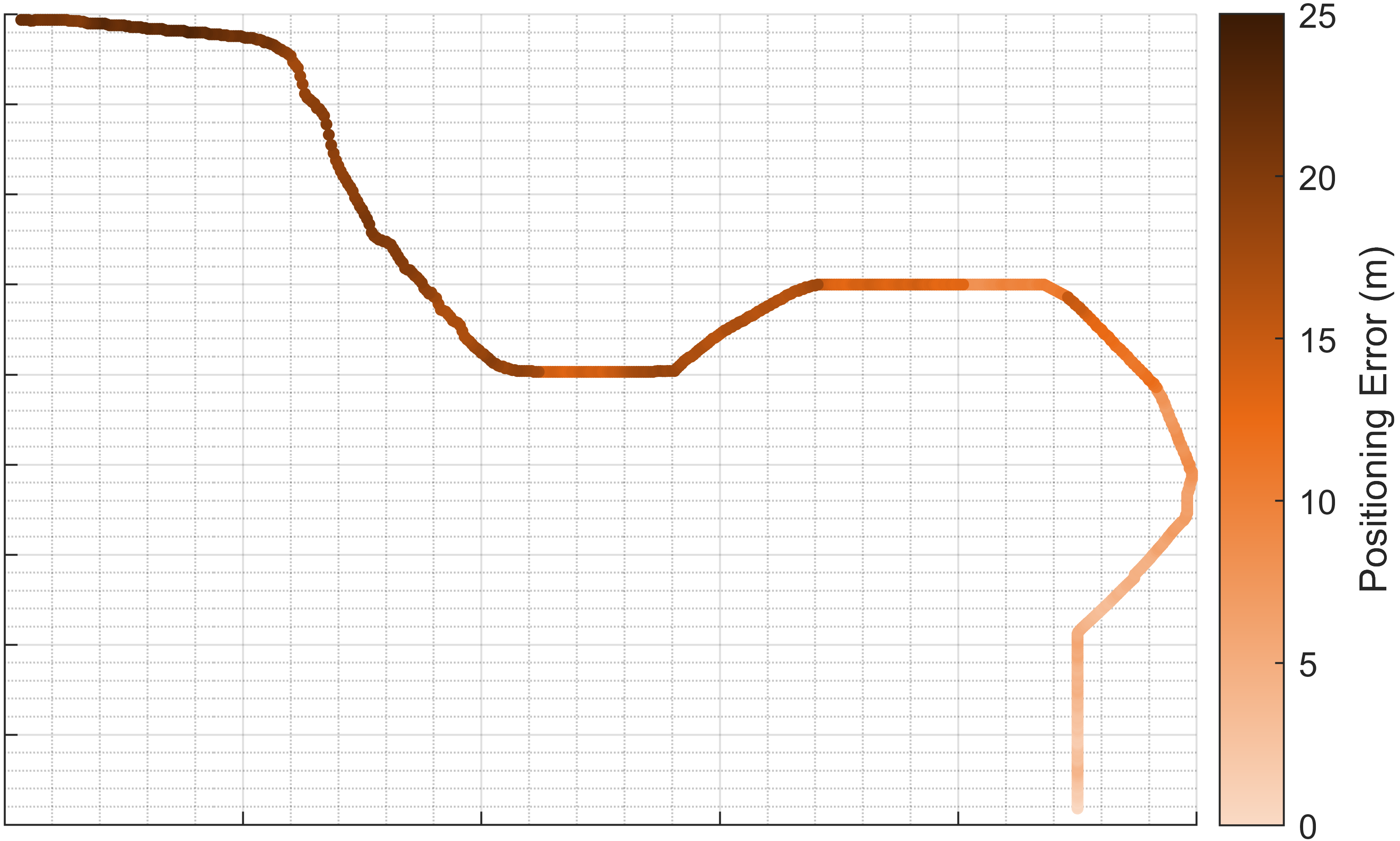}	
			}
		}\noindent
		\subfigure[Ours]{
			\includegraphics[width=3.5in,height=2in]{{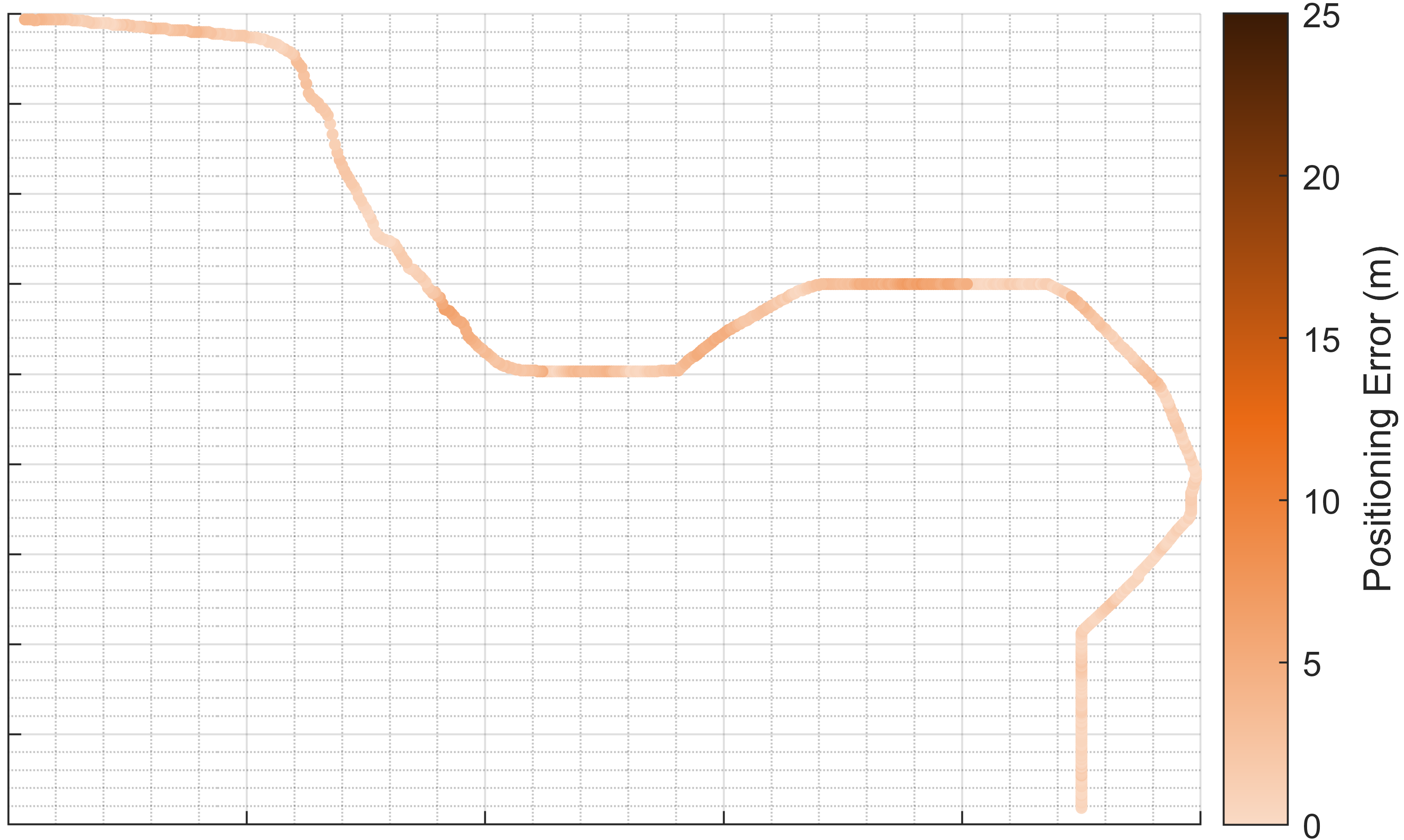}	
			}
		}
	\end{minipage}
        \caption{Illustration of positioning error along the whole route.
        (a) denotes the ground-truth 6.9 km subway route;
        (b), (c) and (d) denote the positioning errors (rendered in colors) of three different methods along the route.}
        \label{fig:fig_7}
\end{figure*}

\subsection{Localization Estimation}
To further analyze and validate the proposed subway self-positioning estimation algorithm, we conduct localization experiments on the VSL Dataset (introduced in Section \ref{sec:dataset}). 
This dataset consists of the 6.9 km driving video data for a subway train,
most of which are running in underground tunnels, providing a favorable environment to validate the effectiveness of the visual-assisted algorithm in tunnel scenarios. 
In addition to the front camera, the train also has a speed sensor to capture the real-time speed, 
including operations such as constant-speed travel, acceleration, and deceleration braking.

\begin{figure}[!t]
\centering
\includegraphics[width=3.4in,height=2.3in]{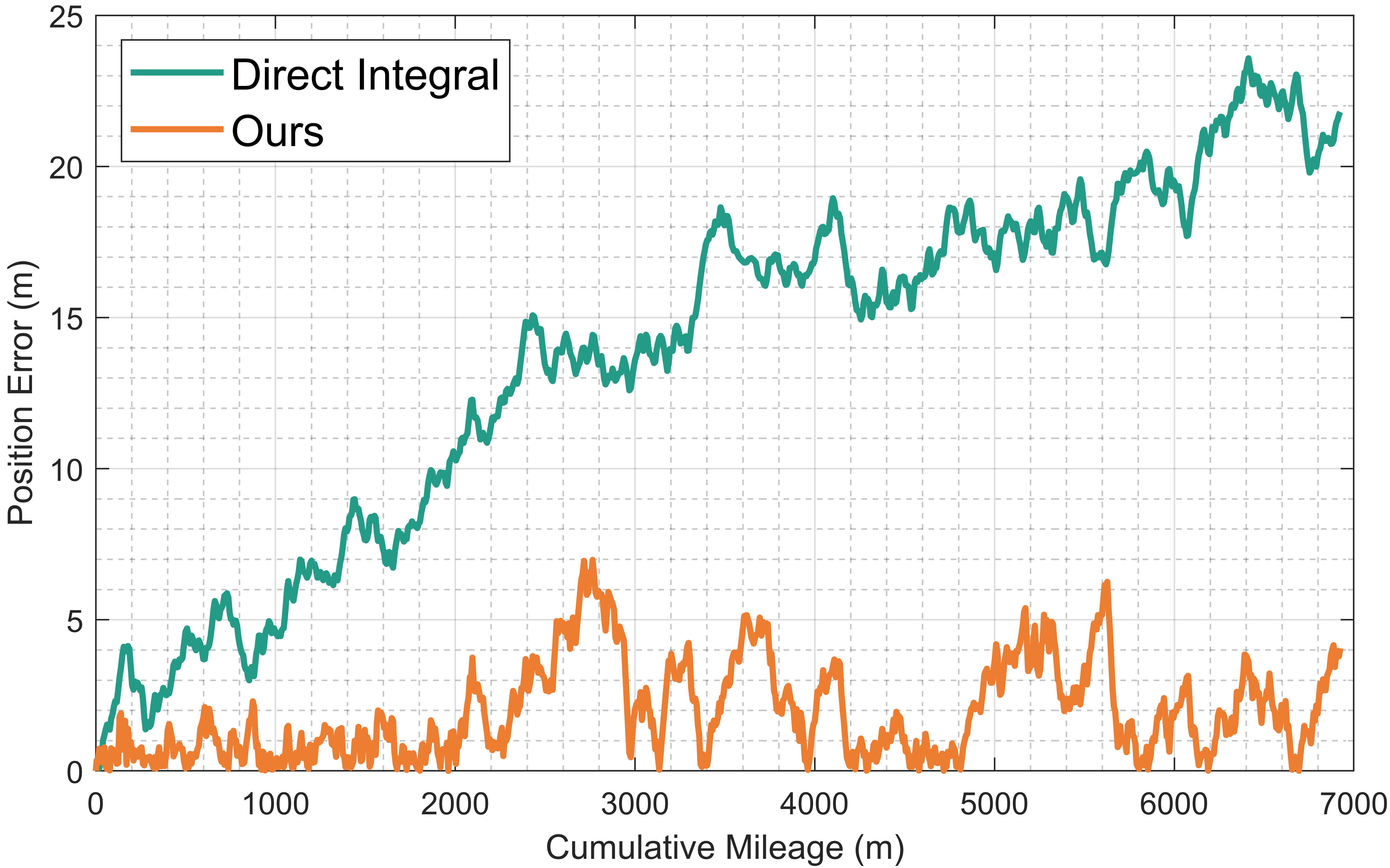}
\caption{Real-time positioning error.}
\label{fig:fig_8}
\end{figure}
The proposed self-localization algorithm mainly relied on the front camera and the speed sensor as input.
It estimates some geometric prior parameters, e.g. $\tau$, $r$, $H$, $d_{B}$ as introduced in Section \ref{sec:module_localization}. 
Compared with direct integration of speed data (Direct Integral) which diverges quickly due to error accumulation,
we show that the proposed method effectively rectifies the time-accumulative errors via the visual correction factors computed with aerial view rail sleeper detection outputs. 
\begin{table}[h]
 \caption{ Inter-Station Localization Errors. \label{tab:table2}} 
 \renewcommand{\arraystretch}{1.6}
\centering
\begin{tabular}{clcc}
\hline
\toprule
Station                 & \multicolumn{1}{c}{Method} & ME & MPE    \\ 
\midrule
  
\multirow{2}{*}{1-2}            & Direct Integral            & 6.28        & 0.93\% \\
                                & Ours                       & 2.30        & 0.28\% \\
\multirow{2}{*}{2-3}            & Direct Integral            & 15.07       & 3.15\% \\
                                & Ours                       & 3.78         & 0.33\% \\
\multirow{2}{*}{3-4}            & Direct Integral            & 18.65        & 5.53\% \\
                                & Ours                       & 6.98        & 1.40\% \\
\multirow{2}{*}{4-5}            & Direct Integral            & 18.95       & 7.55\% \\
                                & Ours                       & 5.38        & 0.98\% \\
\multirow{2}{*}{5-6}            & Direct Integral            & 23.58       & 7.08\% \\
                                & Ours                       & 6.26        & 1.21\% \\ 
\multirow{2}{*}{Whole Route} & Direct Integral            & 23.58       & 0.50\% \\
                & \textbf{Ours}                     &  \textbf{6.98}         & \textbf{0.10\%} \\ 
 \bottomrule
\end{tabular}

\end{table}
As illustrated in Fig. \ref{fig:fig_8} which shows the real-time localization error along the whole 6.9 km subway route,
We can see that the Direct Integral method exhibits an overall increasing trend in positioning error; 
at the end of the route, it reaches a maximum cumulative error of approximately 24 meters. 
In contrast, the proposed system has a fluctuating positioning error, which remains relatively stable. 
For better visualization,
we draw the real-time localization error with different colors in Fig. \ref{fig:fig_7},
which presents a more obvious comparison between the Direct Integral Method and our proposed framework.
To compare with general visual-based positioning methods, 
we also apply the widely recognized visual-based localization method ORB-SLAM3 \cite{R55} on the VSL dataset and show its performance in Fig. \ref{fig:fig_7}.
Consistent with the analysis in Section \ref{sec:ref_self},
this general SLAM method rapidly accumulates irrecoverable errors within a few seconds after the subway enters the tunnel area (denoted by the dashed black line in Fig. \ref{fig:fig_7} (b)). 
This is attributed to the fact that the camera captures adjacent frames with high similarity in tunnel scenes, lacking enough distinctive texture information to provide visual correlations in ORB-SLAM3.
Moreover, the inconsistent illumination conditions
further complicates the feature extraction and matching process.
As a result, 
most of the general visual localization methods like ORB-SLAM3 fail to detect a sufficient number of texture features for matching and positioning in subway scenes.

For a more specific comparison,
we evaluate the inter-station localization error (five intervals between six stations) as well as the overall error in Table \ref{tab:table2}.
Both ME and MPS scores show that the proposed method is consistently better than the Direct Integral algorithm,
achieving an overall MPE of 0.1\%.
We can see that the maximum error (ME) of the Direct Integral method increases proportionally with the overall mileage due to the accumulation of errors, and its mean percentage error (MPE) to the mileage also exhibits an overall rising trend. In contrast, the proposed method demonstrates a relatively smooth overall trend, characterized by smaller fluctuation amplitudes. Notably, the MPE and ME do not exhibit obvious increases over the last few stops. Within specific station intervals, the proposed approach shows a 2 to 4 times decrease in ME and a 3 to 10 times decrease in MPE compared to the Direct Integral Method. Over the entire 6.9 km route, the proposed method successfully reduced the ME error from 23.58 m to 6.98 m, and decreased MPE from 0.5\% to 0.10\%
Another interesting finding is that for the routes totally underground (station intervals 2-3, 4-5, 5-6), the ME and MPE are of the same level as those partially outside of tunnels (station intervals 1-2, 3-4).
This is because the proposed method relies on accurate aerial view sleeper detection results, and is more robust to environmental disturbances compared with feature matching based methods.
Therefore, the proposed method naturally guarantees strong stabilization and robustness.

\section{CONCLUSION}
This paper introduces a novel visual-assisted and cost-effective subway self-localization algorithm. 
The proposed framework comprises an aerial view rail sleeper detection module and a real-time localization estimation module. 
The aerial view rail sleeper detection module conducts visual transformation on the frontal view captured by a monocular camera, and efficiently detects rail sleepers 
Leveraging a convolutional detection network based on YOLOv8n;
while the real-time localization estimation module
achieves accurate location estimation based on the calculated visual correction factors via detected sleepers and geometric constraints. 
The proposed positioning method can rectify accumulated errors in speed sensors, thereby providing the precise positions of subways in textureless underground scenes.
A Vision-based Subway Localization Dataset is collected and annotated for validation of the proposed method,
where we demonstrate the high precision and real-time processing ability of our subway self-localization framework, i.e. achieves the F1 score of 92.9\% in sleeper detection, and can successfully reduce the maximum localization error from 23.58m to 6.98m and decreased the mean percentage error from 0.5\% to 0.1\% compared with the direct speed integral method.
In particular, 
the proposed method has features of strong stabilization and robustness for challenging scenes,  
as its performances are at a similar level for subway running in and outside of tunnels.
%

 \section*{Acknowledgments}
This paper in supported by the Beijing Natural Science Foundation (Grant No.L211014).

\bibliographystyle{IEEEtran}
\bibliography{refer}

\begin{thebibliography}{10}
\providecommand{\url}[1]{#1}
\csname url@samestyle\endcsname
\providecommand{\newblock}{\relax}
\providecommand{\bibinfo}[2]{#2}
\providecommand{\BIBentrySTDinterwordspacing}{\spaceskip=0pt\relax}
\providecommand{\BIBentryALTinterwordstretchfactor}{4}
\providecommand{\BIBentryALTinterwordspacing}{\spaceskip=\fontdimen2\font plus
\BIBentryALTinterwordstretchfactor\fontdimen3\font minus \fontdimen4\font\relax}
\providecommand{\BIBforeignlanguage}[2]{{%
\expandafter\ifx\csname l@#1\endcsname\relax
\typeout{** WARNING: IEEEtran.bst: No hyphenation pattern has been}%
\typeout{** loaded for the language `#1'. Using the pattern for}%
\typeout{** the default language instead.}%
\else
\language=\csname l@#1\endcsname
\fi
#2}}
\providecommand{\BIBdecl}{\relax}
\BIBdecl

\bibitem{R1}
E.~Dimitrova and S.~Tomov, ``Algorithm for positioning of metro trains under communications-based train control,'' \emph{2020 12th Electrical Engineering Faculty Conference (BulEF)}, pp. 1--4, 2020.

\bibitem{R2}
H.~No, J.~Vezinet, and C.~Milner, ``Simplified gnss fusion-based train positioning system and its diagnosis,'' \emph{Proceedings of the 32nd International Technical Meeting of the Satellite Division of The Institute of Navigation (ION GNSS+ 2019)}, 2019.

\bibitem{R3}
N.~Zhu, J.~Marais, D.~B{\'e}taille, and M.~Berbineau, ``Gnss position integrity in urban environments: A review of literature,'' \emph{IEEE Transactions on Intelligent Transportation Systems}, vol.~19, pp. 2762--2778, 2018.

\bibitem{R4}
M.~Yang, G.~Shesheng, and W.~Wei, ``Unscented particle filter based gaussian process regression for imu/bds train integrated positioning,'' \emph{2016 IEEE Information Technology, Networking, Electronic and Automation Control Conference}, pp. 1070--1073, 2016.

\bibitem{R5}
J.~Liu, B.~gen Cai, D.~Lu, and J.~Wang, ``Gps/bds-based virtual balise - enabling satellite-based train control with a train-centric approach,'' \emph{Proceedings of the ION 2019 Pacific PNT Meeting}, 2019.

\bibitem{R6}
O.~Heirich, P.~Robertson, A.~C. Garc{\'i}a, and T.~Strang, ``Bayesian train localization method extended by 3d geometric railway track observations from inertial sensors,'' \emph{2012 15th International Conference on Information Fusion}, pp. 416--423, 2012.

\bibitem{R7}
T.~Albrecht, K.~Luddecke, and J.~Zimmermann, ``A precise and reliable train positioning system and its use for automation of train operation,'' \emph{2013 IEEE International Conference on Intelligent Rail Transportation Proceedings}, pp. 134--139, 2013.

\bibitem{R8}
Z.~Wei, M.~Shengjie, H.~Zexi, J.~Huiwen, and Z.~Zeyu, ``Train integrated positioning method based on gps/ins/rfid,'' \emph{2016 35th Chinese Control Conference (CCC)}, pp. 5858--5862, 2016.

\bibitem{R9}
O.~Heirich, B.~Siebler, and E.~L. Hedberg, ``Study of train-side passive magnetic measurements with applications to train localization,'' \emph{J. Sensors}, vol. 2017, pp. 8\,073\,982:1--8\,073\,982:10, 2017.

\bibitem{R10}
A.~M. Kostrominov, O.~N. Tyulyandin, A.~B. Nikitin, M.~Vasilenko, and A.~T. Osminin, ``Rfid-based navigation of subway trains,'' \emph{2020 IEEE East-West Design \& Test Symposium (EWDTS)}, pp. 1--6, 2020.

\bibitem{R11}
M.-A. Lavoie and J.~R. Forbes, ``Map-aided train navigation with imu measurements,'' \emph{2021 IEEE/RSJ International Conference on Intelligent Robots and Systems (IROS)}, pp. 3465--3470, 2021.

\bibitem{R51}
\BIBentryALTinterwordspacing
J.~Liu, B.~gen Cai, J.~Wang, and D.~Lu, ``Track map–aided capture of virtual balises: A compatible approach to satellite-based railway train control,'' \emph{IEEE Intelligent Transportation Systems Magazine}, vol.~14, pp. 233--249, 2022. [Online]. Available: \url{https://api.semanticscholar.org/CorpusID:226524427}
\BIBentrySTDinterwordspacing

\bibitem{R52}
\BIBentryALTinterwordspacing
J.~Park and S.~H. Cho, ``Fpga implementation for balise telegram decoding and field test validation,'' \emph{IEEE Intelligent Transportation Systems Magazine}, vol.~14, pp. 103--114, 2022. [Online]. Available: \url{https://api.semanticscholar.org/CorpusID:216262084}
\BIBentrySTDinterwordspacing

\bibitem{R12}
T.~Daoust, F.~Pomerleau, and T.~D. Barfoot, ``Light at the end of the tunnel: High-speed lidar-based train localization in challenging underground environments,'' \emph{2016 13th Conference on Computer and Robot Vision (CRV)}, pp. 93--100, 2016.

\bibitem{R37}
J.~Wohlfeil, ``Vision based rail track and switch recognition for self-localization of trains in a rail network,'' \emph{2011 IEEE Intelligent Vehicles Symposium (IV)}, pp. 1025--1030, 2011.

\bibitem{R13}
H.~Huang and S.-K. Yeung, ``360vo: Visual odometry using a single 360 camera,'' \emph{2022 International Conference on Robotics and Automation (ICRA)}, pp. 5594--5600, 2022.

\bibitem{R14}
D.~Scaramuzza and Z.~Zhang, ``Visual-inertial odometry of aerial robots,'' \emph{ArXiv}, vol. abs/1906.03289, 2019.

\bibitem{R15}
R.~C. Smith and P.~C. Cheeseman, ``On the representation and estimation of spatial uncertainty,'' \emph{The International Journal of Robotics Research}, vol.~5, pp. 56 -- 68, 1986.

\bibitem{R16}
J.~P.~M. Covolan, A.~C. Sementille, and S.~R.~R. Sanches, ``A mapping of visual slam algorithms and their applications in augmented reality,'' \emph{2020 22nd Symposium on Virtual and Augmented Reality (SVR)}, pp. 20--29, 2020.

\bibitem{R17}
Y.~Chang, K.~Ebadi, C.~Denniston, M.~F. Ginting, A.~Rosinol, A.~Reinke, M.~Palieri, J.~Shi, A.~Chatterjee, B.~Morrell, A.~akbar Agha-mohammadi, and L.~Carlone, ``Lamp 2.0: A robust multi-robot slam system for operation in challenging large-scale underground environments,'' \emph{IEEE Robotics and Automation Letters}, vol.~7, pp. 9175--9182, 2022.

\bibitem{R18}
Z.~Zhu, J.~Liu, J.~Wang, and N.~Kato, ``Edge-cloud based vehicle slam for autonomous indoor map updating,'' \emph{2020 IEEE 92nd Vehicular Technology Conference (VTC2020-Fall)}, pp. 1--6, 2020.

\bibitem{R19}
F.~Tschopp, T.~Schneider, A.~W. Palmer, N.~Nourani-Vatani, C.~Cadena, R.~Y. Siegwart, and J.~I. Nieto, ``Experimental comparison of visual-aided odometry methods for rail vehicles,'' \emph{IEEE Robotics and Automation Letters}, vol.~4, pp. 1815--1822, 2019.

\bibitem{R20}
Z.~Wang, G.~Yu, B.~Zhou, P.~Wang, and X.~Wu, ``A train positioning method based-on vision and millimeter-wave radar data fusion,'' \emph{IEEE Transactions on Intelligent Transportation Systems}, vol.~23, pp. 4603--4613, 2021.

\bibitem{R53}
\BIBentryALTinterwordspacing
J.~Wang, H.~Zhang, R.~Kang, and P.~Xu, ``An adaptive dynamic coding method for track circuit in a high-speed railway,'' \emph{IEEE Intelligent Transportation Systems Magazine}, vol.~14, pp. 188--199, 2021. [Online]. Available: \url{https://api.semanticscholar.org/CorpusID:233953509}
\BIBentrySTDinterwordspacing

\bibitem{R41}
I.~Watanabe and T.~Takashige, ``Advanced atp system for improving train traffic density and control efficiency,'' \emph{Transportation Research Record}, no. 1314, 1991.

\bibitem{R42}
A.~Hessami, \emph{Modern Railway Engineering}.\hskip 1em plus 0.5em minus 0.4em\relax BoD--Books on Demand, 2018.

\bibitem{R24}
D.~M. Dobkin, ``The rf in rfid, second edition: Uhf rfid in practice,'' 2012.

\bibitem{R25}
A.~Buffi and P.~Nepa, ``An rfid-based technique for train localization with passive tags,'' \emph{2017 IEEE International Conference on RFID (RFID)}, pp. 155--160, 2017.

\bibitem{R26}
Y.~Song and L.~Chen, ``Intelligent localization of a high-speed train using lssvm and the online sparse optimization approach,'' \emph{IEEE Transactions on Intelligent Transportation Systems}, vol.~18, pp. 2071--2084, 2017.

\bibitem{R27}
R.~Cheng, Y.~Song, D.~Chen, and X.~Ma, ``Intelligent positioning approach for high speed trains based on ant colony optimization and machine learning algorithms,'' \emph{IEEE Transactions on Intelligent Transportation Systems}, vol.~20, pp. 3737--3746, 2019.

\bibitem{R43}
\BIBentryALTinterwordspacing
X.~Yali and Z.~Yanpeng, ``A novel method of train positioning based on visible light communication,'' \emph{2020 IEEE 2nd International Conference on Civil Aviation Safety and Information Technology (ICCASIT}, pp. 670--674, 2020. [Online]. Available: \url{https://api.semanticscholar.org/CorpusID:232235543}
\BIBentrySTDinterwordspacing

\bibitem{R31}
K.~Gerlach and C.~Rahmig, ``Multi-hypothesis based map-matching algorithm for precise train positioning,'' \emph{2009 12th International Conference on Information Fusion}, pp. 1363--1369, 2009.

\bibitem{R32}
W.~Nai, Y.~Chen, X.~Zhang, X.~Lei, and D.~Dong, ``A train positioning algorithm based on inflexion analysis by using sets of distance measurement data collected from on-board laser ranging equipments,'' \emph{2017 3rd IEEE International Conference on Computer and Communications (ICCC)}, pp. 901--904, 2017.

\bibitem{R48}
\BIBentryALTinterwordspacing
R.~Liu and C.~Geng, ``Research on train positioning technology of urban rail transit based on lidar,'' \emph{2022 IEEE 7th International Conference on Intelligent Transportation Engineering (ICITE)}, pp. 106--110, 2022. [Online]. Available: \url{https://api.semanticscholar.org/CorpusID:258221104}
\BIBentrySTDinterwordspacing

\bibitem{R33}
C.~Dong, Q.~Mao, D.~Kou, Y.~Dai, Y.~Xiong, X.~Zhu, and X.~Li, ``Dynamic precision surveying of railway tunnels based on mls with robust constraints of track control network,'' \emph{Measurement Science and Technology}, vol.~33, 2022.

\bibitem{R36}
F.~Kaleli and Y.~S. Akgul, ``Vision-based railroad track extraction using dynamic programming,'' \emph{2009 12th International IEEE Conference on Intelligent Transportation Systems}, pp. 1--6, 2009.

\bibitem{R54}
\BIBentryALTinterwordspacing
C.~Li and L.~Zhao, ``A railway turnout closeness monitoring method based on switch gap images,'' \emph{IEEE Intelligent Transportation Systems Magazine}, vol.~14, pp. 214--229, 2022. [Online]. Available: \url{https://api.semanticscholar.org/CorpusID:234334462}
\BIBentrySTDinterwordspacing

\bibitem{R38}
A.~Geiger, P.~Lenz, and R.~Urtasun, ``Are we ready for autonomous driving? the kitti vision benchmark suite,'' \emph{2012 IEEE Conference on Computer Vision and Pattern Recognition}, pp. 3354--3361, 2012.

\bibitem{R44}
\BIBentryALTinterwordspacing
R.~Li, Y.~Lou, W.~Song, Y.~Wang, and Z.~Tu, ``Experimental evaluation of monocular visual–inertial slam methods for freight railways,'' \emph{IEEE Sensors Journal}, vol.~23, pp. 23\,282--23\,293, 2023. [Online]. Available: \url{https://api.semanticscholar.org/CorpusID:260722908}
\BIBentrySTDinterwordspacing

\bibitem{R39}
\BIBentryALTinterwordspacing
A.~Saha, O.~A.~M. Maldonado, C.~Russell, and R.~Bowden, ``Translating images into maps,'' \emph{2022 International Conference on Robotics and Automation (ICRA)}, pp. 9200--9206, 2021. [Online]. Available: \url{https://api.semanticscholar.org/CorpusID:238259883}
\BIBentrySTDinterwordspacing

\bibitem{R57}
\BIBentryALTinterwordspacing
M.~Nieto, L.~Salgado, F.~Jaureguizar, and J.~Cabrera, ``Stabilization of inverse perspective mapping images based on robust vanishing point estimation,'' \emph{2007 IEEE Intelligent Vehicles Symposium}, pp. 315--320, 2007. [Online]. Available: \url{https://api.semanticscholar.org/CorpusID:8226276}
\BIBentrySTDinterwordspacing

\bibitem{R50}
\BIBentryALTinterwordspacing
G.~Jocher, A.~Chaurasia, and J.~Qiu, ``{YOLO by Ultralytics},'' Jan. 2023. [Online]. Available: \url{https://github.com/ultralytics/ultralytics}
\BIBentrySTDinterwordspacing

\bibitem{R58}
M.~Everingham, L.~Van~Gool, C.~K.~I. Williams, J.~Winn, and A.~Zisserman, ``The {PASCAL} {V}isual {O}bject {C}lasses {C}hallenge 2012 {(VOC2012)} {R}esults,'' http://www.pascal-network.org/challenges/VOC/voc2012/workshop/index.html.

\bibitem{R40}
\BIBentryALTinterwordspacing
A.~Bochkovskiy, C.-Y. Wang, and H.-Y.~M. Liao, ``Yolov4: Optimal speed and accuracy of object detection,'' \emph{ArXiv}, vol. abs/2004.10934, 2020. [Online]. Available: \url{https://api.semanticscholar.org/CorpusID:216080778}
\BIBentrySTDinterwordspacing

\bibitem{R45}
\BIBentryALTinterwordspacing
G.~Jocher, ``{YOLOv5 by Ultralytics},'' May 2020. [Online]. Available: \url{https://github.com/ultralytics/yolov5}
\BIBentrySTDinterwordspacing

\bibitem{R49}
\BIBentryALTinterwordspacing
C.-Y. Wang, A.~Bochkovskiy, and H.-Y.~M. Liao, ``Yolov7: Trainable bag-of-freebies sets new state-of-the-art for real-time object detectors,'' \emph{2023 IEEE/CVF Conference on Computer Vision and Pattern Recognition (CVPR)}, pp. 7464--7475, 2022. [Online]. Available: \url{https://api.semanticscholar.org/CorpusID:250311206}
\BIBentrySTDinterwordspacing

\bibitem{R55}
\BIBentryALTinterwordspacing
C.~Campos, R.~Elvira, J.~J.~G. Rodr'iguez, J.~M.~M. Montiel, and J.~D. Tard{\'o}s, ``Orb-slam3: An accurate open-source library for visual, visual–inertial, and multimap slam,'' \emph{IEEE Transactions on Robotics}, vol.~37, pp. 1874--1890, 2020. [Online]. Available: \url{https://api.semanticscholar.org/CorpusID:220713377}
\BIBentrySTDinterwordspacing

\end{thebibliography}

%
\vfill

\end{document}